# A Comparison of the XTAG and CLE Grammars for English

Manny Rayner, Beth Ann Hockey, Frankie James

Research Institute for Advanced Computer Science
Mail Stop 19-39
Moffett Field, CA 94035-1000
USA

## 1. Introduction

When people develop something intended as a large broad-coverage grammar, they usually have a more specific goal in mind. Sometimes this goal is covering a corpus; sometimes the developers have theoretical ideas they wish to investigate; most often, work is driven by a combination of these two main types of goal. What tends to happen after a while is that the community of people working with the grammar starts thinking of some phenomena as "central", and makes serious efforts to deal with them; other phenomena are labelled "marginal", and ignored. Before long, the distinction between "central" and "marginal" becomes so ingrained that it is automatic, and people virtually stop thinking about the "marginal" phenomena. In practice, the only way to bring the marginal things back into focus is to look at what other people are doing and compare it with one's own work.

In this paper, we will take two large grammars, XTAG and CLE, and examine each of them from the other's point of view. We will find in both cases not only that important things are missing, but that the perspective offered by the other grammar suggests simple and practical ways of filling in the holes. It turns out that there is a pleasing symmetry to the picture. XTAG has a very good treatment of complement structure, which the CLE to some extent lacks; conversely, the CLE offers a powerful and general account of adjuncts, which the XTAG grammar does not fully duplicate. If we examine the way in which each grammar does the thing it is good at, we find that the relevant methods are quite easy to port to the other framework, and in fact only involve generalization and systematization of existing mechanisms.

The paper is structured as follows. Section 2 presents a very brief overview of the CLE and XTAG grammars. In Section 3, we describe the CLE grammar from the XTAG grammar's point of view, following which Section 4 describes the XTAG grammar from a CLE perspective. Section 5 concludes.

## 2. An Overview of the XTAG and CLE Grammars

The CLE and XTAG grammars for English are extensively described elsewhere (Pulman, 1992; The XTAG-Group, 1995), and this section will only present the briefest possible summary. Both grammars make a serious attempt to cover all major syntactic phenomena of the language; the CLE grammar also associates each syntactic construction with a compositional scope-free semantics expressed in Quasi Logical Form notation (van Eijck & Alshawi, 1992). In particular, both grammars provide good coverage of the following:

**NP structure:** Pre- and post-nominal adjectival modification, postnominal modification by PPs, relative clauses, -ing and -ed VPs, comparative and superlative adjectives, possessives, complex determiners, compound nominals, time, date and code expressions, numbers, "kind of" NPs, determiner and NBAR ellipsis, sentential NPs, apposition, conjunction of NP.

**Clausal structure:** A large variety of verb types, including intransitives, transitives, ditransitives, copula, auxiliaries, modals, verbs subcategorizing for PPs, particles, embedded clauses, raising and small clause constructions, and combinations of the above; VP modification by PPs, verbal ADVPs, -ing VP, "to" VP declaratives, imperative, WH-questions and Y-N questions; clefts; passives; sentential ADVPs; topicalization; negation; embedded questions; relative clauses; conjunction.

## 3. What XTAG Tells Us About the CLE Grammar

Both grammars are explicitly lexicalized in a way that makes it easy to define a wide variety of types of complement structure. The XTAG grammar defines complement structure through the very flexible and general mechanism of initial trees combined with the adjunction operation for introducing recursion. Very briefly, each initial tree defines one possible complement structure for its head. Complements can be specified as substitution nodes, with features constraining the possible constituents that can be substituted in; alternately, they can be specified as adjunction nodes, which allow auxiliary trees to be adjoined onto them.

CLE grammar, in contrast, defines complement structure through rule schemas. For example, the VP rule schema is of the form

$$VP \rightarrow V:[subcat=COMPS] \mid COMPS$$

the right hand side of which can be glossed as "V whose <subcat> feature has value COMPS, followed by a list of constituents which unify with COMPS". From a TAG perspective, COMPS is more or less equivalent to a list of substitution nodes; there is nothing corresponding to adjunction nodes. The CLE grammar can get along without the adjunction operation, which is absolutely central to XTAG, because it has a powerful mechanism for handling long-range dependencies based on the idea of "gap-threading" (Pereira, 1981; Karttunen, 1986; Pulman, 1992). From the XTAG point of view, it is none the less hard to believe that substitution nodes on their own will be capable of modeling an equally broad range of complement structures.

It does indeed appear to be the case that certain types of complements, particularly those related to idioms and light verbs, are difficult to capture in the CLE framework whereas there is an obvious way to treat these in XTAG. The most convincing example we have identified so far is the class of constructions, very common in English, involving a combination of a verb, a possessive, and a noun, for instance *shake one's head*, *close one's eyes*, *shrug one's shoulders*, *take one's time*, *have one's way*. In all of these constructions, the NP's determiner must be a possessive pronoun agreeing with the verb, and it is in general possible to modify the NP (*shake his pretty head*, *shrug her powerful shoulders*, *have his silly way*). It is obvious that *take one's time* and *have one's way* should be treated as light verb constructions and there are good arguments for modeling the less obvious cases such as *shake one's head*, *close one's eyes* and *shrug one's shoulders* as idioms or light verbs as well, rather than just viewing them as instances of the general transitive verbs shake, close or shrug. For instance, modeling them as idioms or light verbs would be an advantage in the context of a transfer-based machine translation system. Few languages express these concepts in the same way as English[1] and a straight forward compositional treatment will lead to serious complications in defining the associated transfer rules.

---

[1]for example, *close one's eyes* is *fermer lex yeux* in French (transitive verb + definite NP) and *blunda* in Swedish (intransitive verb)

Coding the constraints needed to capture these constructions as idioms is unproblematic in XTAG: for example the initial tree for have one's way will be roughly of the form

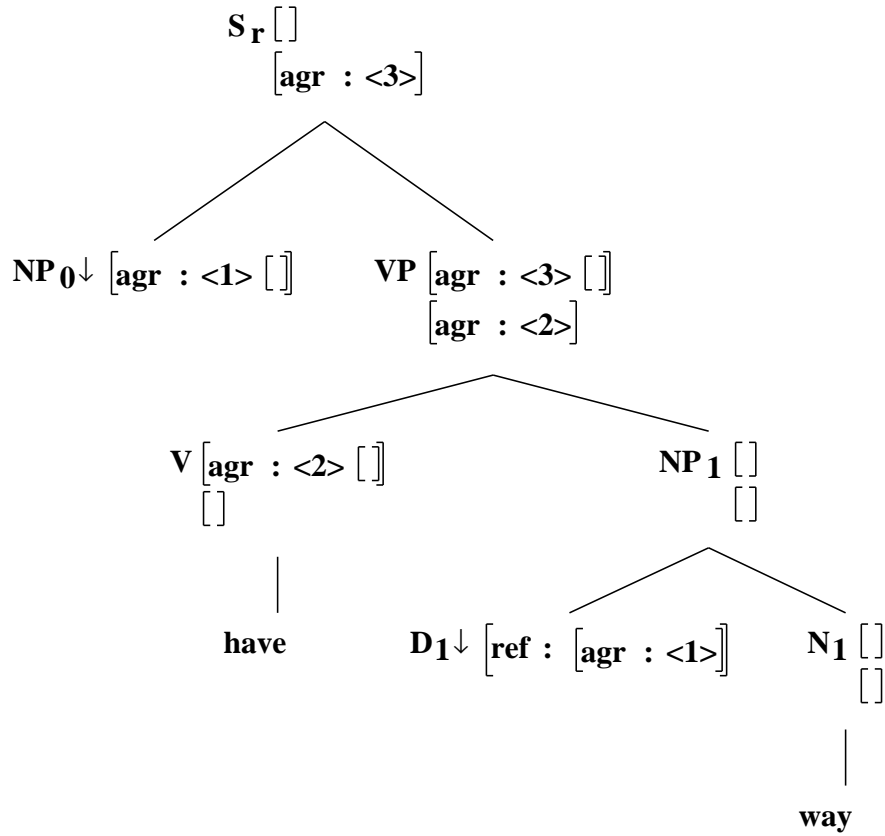

Figure 1: Initial Tree for have one's way

In the current XTAG grammar there is no possessive feature per se. In Figure 1 the determiner is forced to be a possessive pronoun by constraining node D1's <ref> feature to have the same <agr> values as the NP0 and V. Since only pronominal determiners have the <ref> feature, constraining it ensures that the determiner is both pronominal and agrees with the NP0 subject. Notice that because the determiner and the noun of the NP complement are leaves of the tree, it is trivial to state constraints on either of them.

The XTAG treatment cannot be duplicated directly in the CLE framework, since the constraints present in the value of the <subcat> feature are unable to directly reference the DET and N nodes in the complement NP; they can only access that NP's maximal projection. This means that the features on NPs must be such that the relevant information is percolated up through all NP modification rules. Concretely, the category NP needs a head feature which not only specifies whether the DET is a possessive, but also provides agreement information for that possessive; there is however no such feature in the current CLE grammar. We will return to this point in the final section.

## 4.   What the CLE Tells Us About the XTAG Grammar

We now switch to looking at the XTAG grammar from the CLE's point of view. Perhaps the main strength of the CLE grammar is its handling of long-range dependencies, which as al-ready noted is implemented using a gap-threading method. XTAG's main tools for dealing with long-range dependencies are the ability to state constraints within an elementary tree and the adjunction operation. This works very well for some things, in particular most phenomena involving movement of complements; the basic idea is to encode the movement in a suitable initial tree, and let adjunction take care of the rest. None the less, for someone used to the CLE's design philosophy, it is intuitively implausible that all movement phenomena can be captured in this way, and one expects problems with movement of adjuncts. Once again, we looked for a paradigmatic example of the problem; this time, the most clear-cut case appears to be preposition stranding in adjuncts, as illustrated in sentences like which lake did you swim in? The CLE's treatment is fairly straight forward. The sentence receives the phrase-structure

(1) [S [NP which lake]$_j$ [S did$_i$ you [VP [V t$_i$ ] [VP swim] [PP in [NP t$_j$ ]]]]]

in which the empty V constituent is linked to the inverted main verb did, and the empty NP node to the fronted WH+ NP which lake. Features are used to define both kinds of movement. The V is moved through the VP feature <sai> (subject-auxiliary inversion) down to the V gap in the main VP. The NP is moved further, using a gap-threading mechanism, successively through the inner S, the VP, and the PP, to end up coindexed with the NP gap. The mechanisms are described in more detail in (Pulman, 1992).

If we compare the CLE account with that provided by the XTAG grammar, an interesting point emerges. XTAG's treatment of inversion uses the notion of "multi-component adjunction" which is implemented by a feature mechanism. This feature mechanism, described in detail in (Hockey & Srinivas, 1993), forces two elementary trees to act as a "tree set'" by creating a feature clash with the adjunction of the first tree that is resolved by the adjunction of the second.

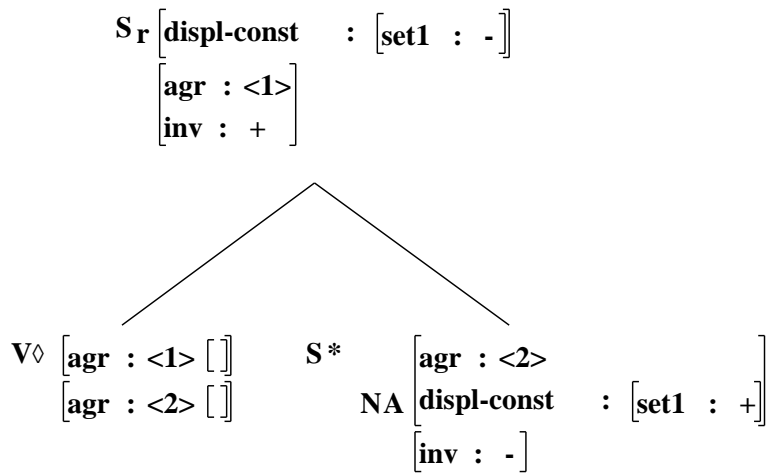

Figure 2: Inverted Verb Auxillary Tree

In the case of inversion the two trees are the tree anchored by the inverted verb shown in Figure 2 and the tree anchored by the verb's trace shown in Figure 3.

The adjunction of either tree individually creates a feature clash between top and bottom feature values of <displ_const> ("displaced constituent") ; however when both trees are adjoined the clash is resolved.

$$\text{VP}_\mathbf{r}\,[\,]$$
$$\begin{bmatrix} \mathbf{agr} : \langle 1 \rangle\,[\,] \\ \mathbf{displ\text{-}const} : \begin{bmatrix} \mathbf{set1} : \langle 2 \rangle\,[\,] \end{bmatrix} \end{bmatrix}$$

$$\text{V}\begin{bmatrix} \mathbf{agr} : \langle 1 \rangle \\ \mathbf{displ\text{-}const} : \begin{bmatrix} \mathbf{set1} : \langle 2 \rangle \end{bmatrix} \\ [\,] \end{bmatrix}$$
$$\text{VP*}\begin{bmatrix} \mathbf{displ\text{-}const} : \begin{bmatrix} \mathbf{set1} : - \end{bmatrix} \end{bmatrix}$$
$$\text{NA}\,[\,]$$

$$\varepsilon$$

Figure 3: Verb Trace Auxillary Tree

Though formally different, one can see that the methods used by the two grammars to treat subject-verb inversion are essentially the same, and involve passing a feature that licenses the coindexing of the fronted main V and the V gap. This is the only type of adjunct movement permitted by the current version of the XTAG grammar. Since Hockey and Srinivas (1993) actually described how the same treatment could easily be used to account for other types of movement, we need to consider why this has not in fact been implemented.

The reason why it is not trivial to extend the current treatment to cover other types of movement is that the information passed by the <displ_const> feature is too coarse-grained; it says that a constituent has been moved, but fails to specify either the type of constituent or the type of movement. A minimal extension of the current framework to cover adjunct NP-movement cases would open the door to promiscuous filler-gap associations and the acceptance of such ungrammatical strings as *Can$_i$ they go to $t_i$* in which the inverted verb can associates with the gap in the PP adjunct to ti. It is clearly necessary to constrain the grammar so as to block these and similar incorrect associations of fillers and gaps.

At this point, it is useful to look at the details of the CLE treatment. The CLE grammar uses features to thread gaps, where the representation of the gaps are feature bundles encoding, among other things, the type of constituent being moved. This immediately suggests one refinement to the XTAG account: if a new feature is added which encodes the category of the moved constituent (call it <displ_cat>), then the worst types of incorrect filler-gap associations can be blocked. Unfortunately, this on its own is not enough since we have to take account of the fact that a constituent can contain more than one gap. The CLE grammar addresses this problem by letting the gap features be list-valued.

It is not clear that the CLE approach can be imported directly into XTAG; given the rather different way in which the two grammars thread features, the CLE's list-valued gap-threading mechanism is hard to combine with the TAG adjunction operator, which the CLE grammar lacks. There is however a straight forward solution. Since there are only a very small number of different types of movement in English that can involve adjuncts, it is possible to use a separate set of features to mediate each type. Specifically, we need four sets of features, which respectively cover verb movement, WH-movement and topicalization, tough movement and right extraposition. (It is possible that passivization forms a fifth class (Pulman, 1987)).

There is nothing linguistically odd about the idea of threading different types of movement separately. It is obvious that subject-verb inversion, WH-movement and right extraposition have different constraints and in most cases operate on different types of constituents. In fact the CLE grammar handles subject-verb inversion and WH-movement through different features and does not cover right extraposition. It does however handle WH-movement and tough-movement through the same set of features, so the interesting question is whether these two should be merged. The most complex aspects of the CLE method are motivated by examples of double extractions involving both WH-movement and tough movement. The main consideration is to enforce the no-crossing dependencies (NCD) constraint as illustrated by the well known "sonata sentences" below; we want to allow (2) and block (3).

(2) Which violin$_i$ are these sonatas$_j$ hard to play t$_j$ on t$_i$?
(3) *Which sonatas$_i$ is this violin$_j$ hard to play t$_i$ on t$_j$?

This provides the main justification for using list as opposed to set valued features to implement gap threading (Pulman 1992, pp 71-73). Although a detailed discussion of the NCD constraint is beyond the scope of this paper, it is clear that it applies more strongly to extractions from complements than to extraction from adjuncts[2]. Since the gap threading mechanism would only be used by the XTAG grammar for adjuncts, the critical examples are those that involve double extraction from adjunct positions. Examples of this kind are first of all very rare, and it is not at all clear that the NCD constraint holds for them. For instance example (4) which breaks the constraint seems if anything more natural than the version with no crossing extractions in (5)

(4) Which articles$_i$ are men$_j$ most fun to shop for t$_i$ with t$_j$?
(5) Which articles$_i$ are men$_j$ most fun to shop with t$_j$ for t$_i$?

To sum up, it seems fair to say that the idea of using separate features to thread WH-movement gaps and tough movement gaps is at least no worse than the CLE's list-valued scheme, which merges them into a single set of features. Our overall conclusion is that the treatment we have sketched above represents a fully viable approach to adapting the CLE gap threading treatment to the problem of handling adjunct extractions in XTAG.

## 5. Summary and Conclusions

Looking at the examples in Sections 3 and 4, we see a common pattern. In each case, one grammar can do the job; the other one almost gets there, but falls over at the last moment. Intuitively, one feels that the problem is in neither case impossible to solve.

Let us first look at the example with have one's way from Section 3. As noted, the CLE could deal with this kind of construction if NPs just had the right head features. The reason these features aren't present is not particularly deep; no one saw a need for them, so they were never put in. Since they have to be trailed through a large number of rules involving NPs, the effort needed to add them is non-trivial, and without a concrete reason to attack the problem things stayed as they were. It would however be quite easy, in principle, to make a careful study of the types of features needed to cover the constructions which the XTAG grammar can deal with. If all of these features were added at once, using sensible macro invocations to do the threading, the work required would not in fact be very frightening. What is more, properly designed macros would make it easy to add new head features as and when they were found to be necessary. The biggest step to take is noting that there is a problem, and making the decision to do something about it.

---

[2]We would like to thank Bob Levine for insightful discussion on this and other points relating to the NCD constraint.

The difficulties involving movement of adjuncts discussed in Section 4 are less trivial, but nonetheless quite soluble. Though these problems have been recognized for some time and suggestions made about how to provide the necessary additional constraint in the XTAG grammar, a system for doing this has not been implemented. As far as we can see, the real explanation is once again a combination of software engineering considerations and research sociology: a vague feeling that the solution was complex and inelegant, and would involve more effort that would be justified to cover a set of "marginal" phenomena. In actual fact, a comparison with the CLE grammar convinces us that the XTAG group was wrong on all counts. The solution appears fairly principled, and is not very hard to implement; and the phenomena in question, far from being marginal, are at least as central as many of those already covered.

To summarize, we have compared the CLE and XTAG grammars, and found some important and non-trivial problems. The CLE is unable to duplicate some of the complement structure phenomena handled by XTAG, and this appears to be due to an insufficiently detailed modeling of head features. Conversely, XTAG is unable to encode some types of constructs involving adjuncts and movement, and we have suggested that the CLE's gap-threading treatment could be adapted to a implement a more general version of multi-component adjunction. However, we think the real moral of the paper is much more fundamental: if people developing big grammars want to make serious progress, it would be in everyone's interest to carry out this kind of detailed comparison more regularly! We hope that our remarks will encourage other researchers to do so.